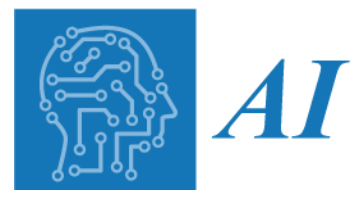

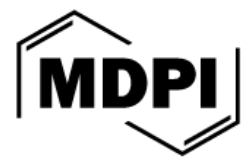

*Article*

# FHIR-RAG-MEDS: Integrating HL7 FHIR with Retrieval-Augmented Large Language Models for Enhanced Medical Decision Support

**Yildiray KABAK [1,*], Gokce B. LALECI ERTURKMEN [1,*], Mert GENCTURK [1], Tuncay NAMLI [1], A. Anil SINACI [1], Ruben ALCANTUD CORCOLES [2,3], Cristina GOMEZ BALLESTEROS [2,3], Pedro ABIZANDA [2,3,4], Asuman DOGAC [1]**

[1] SRDC Software Research & Development and Consultancy Corporation, Ankara, Türkiye
[2] Geriatrics Department, Complejo Hospitalario Universitario de Albacete, Albacete, Spain
[3] CIBER de Fragilidad y Envejecimiento Saludable (CIBERFES), Instituto de Salud Carlos III, Madrid, Spain
[4] Facultad de Medicina de Albacete, Universidad de Castilla-La Mancha, Albacete, Spain
* Correspondence: yildiray@srdc.com.tr

**Abstract**

Evidence-based clinical guidelines are essential for high-quality care yet translating them into personalized clinical decision support remains resource-intensive and time-consuming. Large language models (LLMs) show promise for supporting clinical decision-making, but their limited access to patient-specific data and explicit guideline sources constrains trustworthiness, personalization, and clinical applicability. Retrieval-augmented generation (RAG) addresses part of this challenge by grounding model outputs in curated evidence sources; however, true personalization requires structured access to electronic health record data. This study presents FHIR-RAG-MEDS, a medical decision support system that integrates HL7 Fast Healthcare Interoperability Resources (FHIR) with a RAG-enhanced LLM to enable patient-specific, guideline-concordant clinical recommendations. Through SMART on FHIR, the system retrieves real-time patient data from FHIR servers, generates structured medical summaries, and incorporates this personalized context into the RAG pipeline, grounding responses in evidence-based clinical guidelines stored in a vector database. FHIR-RAG-MEDS was evaluated using 70 physician-generated clinical questions covering dementia, chronic obstructive pulmonary disease, hypertension, and sarcopenia. Performance was assessed using automated metrics, RAG-specific evaluation frameworks, and independent expert physician review. The system consistently outperformed state-of-the-art medical LLMs, demonstrating higher semantic accuracy, improved faithfulness to guideline content, and stronger clinical relevance. Integrating HL7 FHIR with RAG-based LLMs enables trustworthy, personalized clinical decision support, bridging the gap between static language models and real-world, patient-centered care.

## 1. Introduction

Evidence-based clinical guidelines serve as vital tools to support healthcare professionals in delivering high-quality care [1]. Yet, the manual application of these guidelines, especially in complex multimorbidity scenarios, often leads to inconsistent practices, inefficiencies, and suboptimal outcomes, undermining the goals of integrated care [2,3]. In this context, computer-interpretable clinical guidelines are critical for digitizing healthcare and enabling the implementation of personalized clinical decision support (CDS) systems [4]. These systems can integrate patient-specific data with guideline recommendations, providing real-time, tailored support to care teams.

Recent studies such as C3-Cloud [5], ADLIFE [6], HYP [7] and CAREPATH [8] have pioneered integrated care solutions based on CDS systems that automate evidence-based guidelines. These projects have demonstrated a reliable mechanism for translating clinical guidelines into personalized care plans, involving selection of international guidelines by clinical reference groups, technical analysis and specification of CDS requirements [8], and implementation of CDS services to support healthcare professionals in real-world practice [7]. However, these processes are resource-intensive, often requiring 3–6 months for guideline analysis and stakeholder engagement, followed by another 3–6 months for CDS development, testing, and validation.

In recent years, the field of medical informatics has seen significant advancements with the introduction of medical large language models (LLMs). These models, powered by artificial intelligence, have demonstrated remarkable capabilities in understanding and generating medical text, providing valuable assistance in clinical decision-making, diagnostics, and patient care. Prominent examples include models such as Meditron [9], BioMistral [10] and OpenBioLLM [11], which have shown considerable promise in various medical applications. However, despite these advancements, the inherent limitations of medical LLMs highlight the need for more robust solutions.

While medical LLMs have achieved impressive results, they are not without limitations. These models primarily rely on their training data and do not have inherent mechanisms to incorporate real-time patient-specific information or to update their knowledge base dynamically [12] especially to focus on a specific context. As a result, their recommendations may lack contextual relevance and may not fully account for the nuances of individual patient cases, recent developments in medical guidelines or customized approaches in local guidelines. This can lead to less accurate or suboptimal recommendations in a clinical setting. Retrieval-Augmented Generation (RAG) represents a significant advancement over traditional LLM approaches by addressing some of these limitations, by combining the strengths of retrieval-based methods with generative capabilities, enabling the system to access and incorporate specific, up-to-date information from a curated knowledge base [13]. This approach allows the RAG system to generate recommendations that are not only based on broad training data but also tailored to the context of the patient's medical history and current/local clinical guidelines. The main advancements that RAG systems offer over medical LLMs are as follows.

1. Accuracy and Trustworthiness: Medical LLMs rely solely on static training data, which may become outdated as medical knowledge evolves [14], increasing the risk of incorrect or misleading outputs [15]. By retrieving content from validated clinical guidelines, databases, and scientific literature, RAG systems ground generated responses in up-to-date and trustworthy evidence, significantly reducing hallucinations [16,17]. This capability is particularly important for aligning recommendations with national and regional clinical guidelines, thereby improving compliance and reducing legal risk.

2. Evidence-Based Responses: Medical LLMs typically generate recommendations without explicit source attribution, limiting transparency and clinician trust [18]. RAG systems combine generative AI with retrieved evidence, enabling responses that are explicitly grounded in and traceable to clinical sources, a critical requirement for clinical validity and acceptance in healthcare settings [19].
3. Scalability and Flexibility: Training medical LLMs for each domain is resource-intensive and inefficient for cross-domain use. RAG systems instead retrieve relevant information from specialized knowledge bases, allowing efficient support across multiple medical domains and guideline sets without retraining.
4. Reduced Computational Costs: Medical LLMs require heavy computational resources to generate responses, particularly as model sizes increase, leading to slower response times and higher infrastructure costs [20]. RAG systems improve efficiency by retrieving only relevant information prior to generation, reducing inference time and infrastructure requirements.

The integration of RAG with LLMs in the medical domain is still in its early stages [19]. Additionally, to fully harness the potential of LLMs and RAG in routine medical practice, they need to access historical Electronic Health Record (EHR) data. This access enables LLMs to enhance clinical decision support by offering more personalized and contextually relevant recommendations based on a patient's medical history. Studies such as [21,22] have reviewed the integration of LLMs with EHRs, outlining both the potential and limitations of this approach. They point out that while LLMs can assist with tasks such as diagnostic support and treatment planning, the challenge lies in effectively incorporating patient data into these models.

This study aimed to evaluate a RAG enhanced LLM system that is directly integrated with EHRs, to assess its effectiveness in providing personalized trustworthy decision support in accordance to evidence based clinical guidelines. We address the gap of personalization by integrating Health Level 7 Fast Healthcare Interoperability Resources (HL7 FHIR) interfaces into an RAG system that will act as a co-pilot for medical professionals. By leveraging the standardized data formats provided by HL7 FHIR, our system can retrieve up-to-date patient medical summaries and utilize them to retrieve personalized suggestions from evidence-based clinical guidelines. This ensures more accurate, personalized, and contextually relevant recommendations, overcoming the limitations regarding the static nature of LLMs that lack dynamic access to patient data. Although there are studies on the use of LLMs on HL7 FHIR, they use it to increase patient's health literacy on their medical records [23] and to convert the unstructured medical records into HL7 FHIR resources [23,24]. In our work, HL7 FHIR integration allows our system to provide real-time decision support, enhances interpretability, and enables personalization of recommendations, making it better suited for clinical environments where personalized guidance based on medical standards is critical.

## 2. Related Work

Retrieval-Augmented Generation (RAG) was proposed by [12] to enhance generation performance on knowledge-intensive tasks by integrating retrieved relevant information. RAG not only mitigates the problem of hallucinations, as LLMs are grounded in given contexts, but it can also provide up-to-date knowledge that might not be encoded within the LLMs. In the medical domain, there have already been various explorations into the use of LLMs with RAG. One prominent study highlights the use of RAG in LLMs tailored for specific medical domains, such as nephrology [13]. In this context, RAG allows these models to access real-time external medical databases, including clinical guidelines, to provide contextually relevant and up-to-date information. This approach significantly enhances the model's performance by grounding responses in validated, evidence-based guidelines, which is critical in healthcare settings where precision is vital. For example, this methodology helps the model generate more accurate recommendations for conditions like kidney disease. Another interesting example is the experimental LLM framework, Almanac, which integrates RAG functionalities specifically with clinical guidelines and medical treatments [19]. This framework was tested against existing models, such as ChatGPT, and demonstrated superior performance, particularly in cardiology, where it provided more accurate responses compared to standard generative models. A similar RAG based system [25] has been developed and evaluated for querying the United Kingdom's National Institute for Health and Care Excellence (NICE) clinical guidelines using LLMs. It offers healthcare professionals the ability to obtain rapid, evidence-based answers to clinical questions while offering a flexible, modular framework that can be adapted to any clinical guideline corpus. An additional recent study [26] has evaluated a RAG-enhanced LLM system, grounded in European Association of Urology and American Urological Association guidelines, to assess its effectiveness in providing guideline-concordant Prostate-specific antigen (PSA) screening recommendations compared to junior clinicians. These studies, however, do not focus on EHR integration and personalization of recommendations based on direct processing of the patient's medical summaries recorded in EHRs.

Studies such as [21,22] have reviewed the integration of LLMs with EHRs. The integration is centered around seven identified topics: named entity recognition, information extraction, text similarity, text summarization, text classification, dialogue system, and diagnosis and prediction. However, none of the examined studies use standardized data models to access medical records. A notable patient-centered example was introduced by [23], which combines GPT-4 with HL7 FHIR to enhance the health literacy of a diverse patient population. However, this approach is aimed at patients and does not utilize RAG. Our system, in contrast, is designed specifically for medical professionals and is built on evidence-based clinical guidelines. This ensures that recommendations are not only accurate and trustworthy but also up to date, overcoming the limitations of static, pre-trained models like GPT-4.

In summary, medical RAG systems can improve EHR summarization [27], clinical decision-making [13,19,20,28] and clinical trial screening [29], but their evaluations are not comprehensive [30]. Furthermore, despite promising results, existing medical RAG systems largely operate in isolation from electronic health records and lack seamless integration through standardized interoperability frameworks such as HL7 FHIR. This prevents direct use of patient-specific medical summaries in the decision process, significantly constraining personalization and limiting their suitability for routine, real-world clinical decision support.

## 2. Materials and Methods

This study aims to describe the development and evaluation of the FHIR-RAG-MEDS system that integrates HL7 FHIR with a RAG-based system to improve personalized medical decision support based on evidence-based clinical guidelines.

In Section 2.1, the case study details are presented elaborating on the scope of the evaluation study protocol. In Section 2.2, we present the technical details of the implementation of the RAG-Enabled decision support service, FHIR-RAG-MEDS, including preprocessing of guideline text to create embeddings to populate vector database, the data retrieval and LLM supported medical summary creation and segmentation phase, and RAG execution phase. To ensure reproducibility, the datasets processed (guideline text, sample hypothetical patient data as FHIR resources, medical summary extract) are provided as *Supplementary Materials 1 and 2*. The libraries utilized for text segmentation and their configuration parameters, Vector database used, similarity metrics used, and embedding model used in the pre-processing step are clearly described in Section 2.2.1. The software code to enable integration with a Smart On FHIR enabled patient data repository as presented in Section 2.2.2 is made available as open source [31]. The prompt template we utilize for pre-processing FHIR bundles with Llama 3.1 8B model to convert the JSON into a textual medical summary is made available as *Supplementary Material-3*. RAG Execution results including the question asked, answer received, the corresponding ground truth and the Prometheus evaluation is made available as *Supplementary Material-4*.

In Section 2.3, we present the details of the evaluation framework. To ensure the reproducibility, the 70 question and answer pairs created by the evaluation panel as fictional case scenarios are provided as *Supplementary Material-5*. The details of the automated evaluation tools utilized along with the prompts used are presented in Section 2.3.

Evaluation results are presented in section 3, along with the full list of evaluation questions and physicians' responses which are provided as *Supplementary Material-6*.

### *2.1. Case Study Selection*

As a case study, we have chosen a European Commission (EC)-funded research project, CAREPATH [32] which aims to deliver a patient-centered integrated care platform to meet the needs of older patients with multimorbidity, including mild cognitive impairment (MCI) or mild dementia (MD), based on the recommendations of evidence-based guidelines. In the CAREPATH project, a Clinical Reference Group (CRG), formed by the project's clinical partners, has analyzed several clinical guidelines that address the needs of common comorbidities in this patient group and created a consolidated guideline providing advice, information, and actions for the management of MCI/MD, physical exercise, nutrition and hydration, common use of drugs, coronary artery disease, heart failure, hypertension, diabetes, chronic kidney disease, chronic obstructive pulmonary disease (COPD), stroke, sarcopenia, and frailty [33]. Subsequently, this consolidated guideline was analyzed by software engineers to create clinical decision support services that automate the recommendations from the guideline, delivering personalized suggestions for care plan goals and activities to healthcare professionals [8]. Although this method provides a solid and reliable mechanism to create clinical decision support for healthcare professionals, the process requires significant time and resources. In this research, we aim to evaluate the value of utilizing advanced large language models and the RAG methodology to accelerate the delivery of personalized clinical recommendations to healthcare professionals based on a selected curation of evidence-based guidelines. As a case study, we have focused on a subset of the CAREPATH consolidated guideline, covering recommendations for managing hypertension, COPD, sarcopenia, and MCI/MD.

The CAREPATH study is conducted in accordance with the Declaration of Helsinki and has been approved by the Ethics Committee of Albacete (V4.0 protocol code 2020-31, No-EPA, and date of approval 25/04/2024) for studies involving humans. Informed consent was obtained from the evaluating healthcare professional participants, who are also the co-authors of this submission, involved in the study. This study did not involve real patient data. As a part of the evaluation study, we have tasked our evaluation panel of physicians to generate 70 questions and answer pairs as fictional case scenarios to represent a broad spectrum of clinical presentations of dementia, chronic obstructive pulmonary disease, hypertension, and sarcopenia patients regarded as ground truths.

### *2.2. System Architecture*

The FHIR-RAG-MEDS system architecture (Figure 1) integrates HL7 FHIR with a RAG framework to deliver real-time, evidence-based clinical decision support. The architecture is designed to retrieve patient-specific data from an HL7 FHIR server, process clinical guidelines stored in a vector database, and generate recommendations using LLMs. The system flow is divided into three core components, as depicted in Figure 1: preprocessing, data retrieval & query processing, and RAG execution:

- **Preprocessing**: This step is required to prepare the guideline text to be processed by the RAG framework. It includes segmenting the text into smaller chunks and generating embeddings from these chunks to populate the vector database, which serves as an input to the RAG system.
- **Data Retrieval and Query Processing**: In this phase, patient context data is retrieved from the FHIR server and pre-processed to create a medical summary, which is integrated into the RAG system as input. SMART on FHIR (Substitutable Medical Applications and Reusable Technologies on Fast Healthcare Interoperability Resources) specifications are employed at this stage. SMART on FHIR is a widely adopted framework that enhances interoperability in healthcare by providing standardized, secure, and scalable solutions for accessing and exchanging clinical data [26]. Unlike proprietary or isolated solutions, SMART on FHIR fosters a robust ecosystem where applications can interact with any FHIR-compliant server, ensuring consistent access to patient data across platforms. This framework is particularly valuable for clinical decision support systems like FHIR-RAG-MEDS, where real-time access to patient-specific data is crucial for generating accurate, evidence-based recommendations. Next, the user's plain text query (e.g., 'What should be the approach to pharmacological treatment for Patient X?') is merged with the generated medical summary text (e.g., 'Patient X is an 85-year-old female patient with cognitive impairment and a history of injurious falls. She is diagnosed with hypertension') and processed to extract embeddings, which are stored in the vector database.
- **RAG Execution**: This step integrates the generative capabilities of LLMs with the retrieval of specific, up-to-date information from the vector database containing preprocessed guidelines. It matches the embeddings extracted from the patient's medical summary and the user query with the guideline embeddings in the vector storage to identify the closest matching vectors. These matching vectors, along with the medical summary, are then used by the LLM to generate a response to the clinician's query.

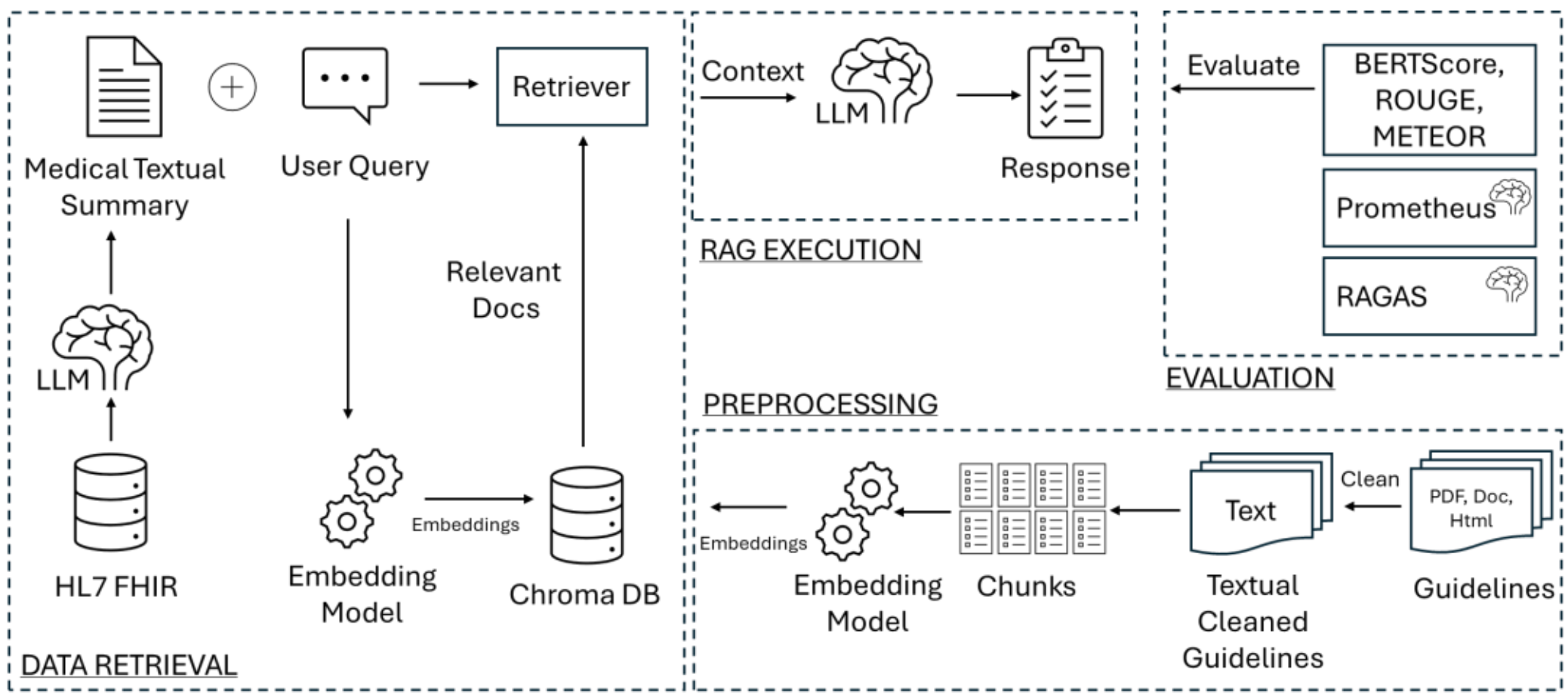

**Figure 1 –** Overall architecture of the FHIR-RAG-MEDS system

In the following subsections, we provide a detailed explanation of these steps, including our design decisions and the tools utilized.

2.2.1. Preprocessing

Handling semi-structured data, such as PDFs containing a mix of text and tables, presents specific challenges for traditional RAG systems. First, text splitting may inadvertently break apart tables, causing data corruption during retrieval. Second, incorporating tables into the data complicates semantic similarity searches [16]. To address this, clinical documents in various formats, including PDFs, should be converted to a text-based format to ensure compatibility with the RAG framework [28].

Once converted, the processed texts are segmented into smaller chunks for embedding and retrieval. Tools from the Langchain library [34] are employed in this splitting process, typically creating chunks of arbitrary sizes with slight overlaps. Determining the optimal chunk size for healthcare applications is a nuanced task that requires qualitative evaluation. In our study, we used Python 3.12 along with Langchain's DirectoryLoader, RecursiveCharacterTextSplitter, and TextLoader for reading text-based clinical guidelines. We established a chunk size of 1200 units, with a 100-unit overlap, based on Ollama's "mxbai-embed-large" embedding model.

Efficient data retrieval also depends on robust data storage solutions. Vector storage, inspired by deep learning techniques, condenses information into high-dimensional vectors, enhancing retrieval performance. The key metrics for evaluating vector storage include cosine similarity, Euclidean distance, and dot product. Cosine similarity is particularly effective for semantic searches and is preferred in healthcare applications because it focuses on the angle between vectors, highlighting content similarity [20]. Conversely, Euclidean distance is better suited for quantitative assessments. Embedding, the process of converting content into numerical vectors for machine learning models, is essential for transforming preprocessed healthcare data into vectors. For this study, we used Chroma [35] as our vector storage solution with cosine similarity, and as previously mentioned, employed Ollama's "mxbai-embed-large" model for embedding.

2.2.2. Data Retrieval and Integration with Existing EHR Systems

The architecture begins with the HL7 FHIR server, which stores structured patient data following the FHIR standard. To enhance interoperability and scalability, a "SMART on FHIR"-based integration has been developed for the FHIR-RAG-MEDS system. This integration leverages the SMART on FHIR framework, which provides a standard set of specifications for building secure, interoperable healthcare applications. These applications can seamlessly connect with any FHIR server that conforms to these standards.

The integration was implemented using Angular, a modern and responsive web development framework, to create a user-friendly interface. This Angular-based application communicates with the FHIR-RAG-MEDS system's endpoints for real-time patient data interpretation and clinical query processing. By adhering to SMART on FHIR specifications, the integration ensures compatibility with various FHIR servers while simplifying authentication, authorization, and data exchange processes.

Key components of the SMART on FHIR integration include the following:

1. Authentication and Authorization: The system employs OAuth 2.0 protocols, as defined by SMART on FHIR, to securely authenticate users and authorize access to patient data. This ensures that sensitive medical information remains protected while maintaining seamless usability.
2. Data Retrieval and Management: The integration enables real-time retrieval of patient information, such as demographics, medications, conditions, and observations, from any compliant FHIR server. All recent condition, medication, and observation resources are retrieved individually and then combined into a FHIR bundle for interpretation by the FHIR-RAG-MEDS system.
3. Interoperable Query Processing: Once patient data is retrieved and summarized, it is sent to the FHIR-RAG-MEDS system for interpretation and recommendation generation as explained in the next section. The system's query endpoints provide evidence-based clinical suggestions tailored to the patient's specific medical context.
4. Extensibility and Compatibility: The SMART on FHIR integration is designed to be extensible, allowing it to support additional use cases, such as integration with third-party applications, EHR systems, and telehealth platforms. By adhering to open standards, it ensures compatibility across diverse healthcare ecosystems.

#### 2.2.3. Processing FHIR Bundles

When a query is initiated to retrieve recommendations from evidence-based guidelines for a specific patient, relevant patient information, such as demographics, medical history, medications, and conditions, is retrieved from the FHIR server once Smart on FHIR based integrating is achieved. This retrieval is performed through FHIR-compliant queries using the appropriate APIs. By leveraging standardized FHIR resources, the system ensures interoperability and consistency across different healthcare environments. Once the patient data is retrieved in FHIR JSON format, it must be processed and organized into a format suitable for integration with the RAG system. For this, we use the Llama 3.1 8B [36] model with the prompt template (see *Supplementary Material-3*) to convert the JSON into a textual medical summary, enabling its inclusion in the LLM context.

The prompt ensures that the retrieved data is transformed into a concise and clear summary, ready for interpretation by the system without overwhelming the user with technical jargon. By filtering and summarizing the patient's medical information in this way, the system enhances usability while maintaining the factual accuracy and relevance necessary for medical decision-making.

The Llama 3.1 8B model has been selected for interpreting FHIR bundles based on several key factors. Testing with an example FHIR bundle set has demonstrated that the model performs well in generating accurate and concise medical summaries. Its relatively small size, up-to-date architecture, and popularity within the community further supported its selection. As an open-source model, Llama 3.1 8B is also easily deployable, even in resource-limited environments such as laptops. Furthermore, the ability to run the model within a local network improves security and privacy, ensuring that confidential health data remains protected without relying on external cloud services.

Sample FHIR resources and medical summaries generated by the system are presented as *Supplementary Material-2*.

2.2.4. RAG Execution

The core of the system lies in the RAG module, which combines the generative capabilities of LLMs with the retrieval of specific, up-to-date information from a pre-processed knowledge base. Clinical guidelines, which are cleaned and stored in a vector database, are indexed to enable fast retrieval of evidence-based recommendations in the previous step.

The RAG execution functions as an intermediary, identifying the most relevant chunks in response to user queries. This involves using the same embedding model to convert user queries into vectors, which are then filtered through the Vector Storage to locate the closest matching vectors. The number of chunks retrieved, adjustable through a parameter commonly referred to as 'k', can be set to any desired value. In our implementation, we set 'k' to 4, determining the number of knowledge chunks retrieved. After that, we combined these chunks with the FHIR Bundle textual interpretation generated in the previous step and created a context to be sent as an input to the LLM. The system uses Llama 3.1 8B, installed via Ollama on a local server, to generate responses to clinician queries.

The integration of FHIR with RAG allows the system to generate personalized clinical recommendations by combining static, evidence-based guidelines with dynamic, real-time patient data. This architecture ensures that the recommendations are both contextually relevant to the individual patient's health profile and aligned with the latest medical standards. As a part of *Supplementary Material-4*, we present the RAG Execution results including the question asked, answer received, the corresponding ground truth and the Prometheus evaluation.

We have created a proof-of-concept user interface to run the FHIR-RAG-MEDS as a co-pilot for the clinicians. As shown in Figure 2, in the left pane, a summary of the patient data retrieved from the FHIR repository is summarized. The medical summary as text is automatically presented in the chat interface. A conversational interface is enabled in the chatbox, where the clinician can ask questions and receive guideline based recommendations.

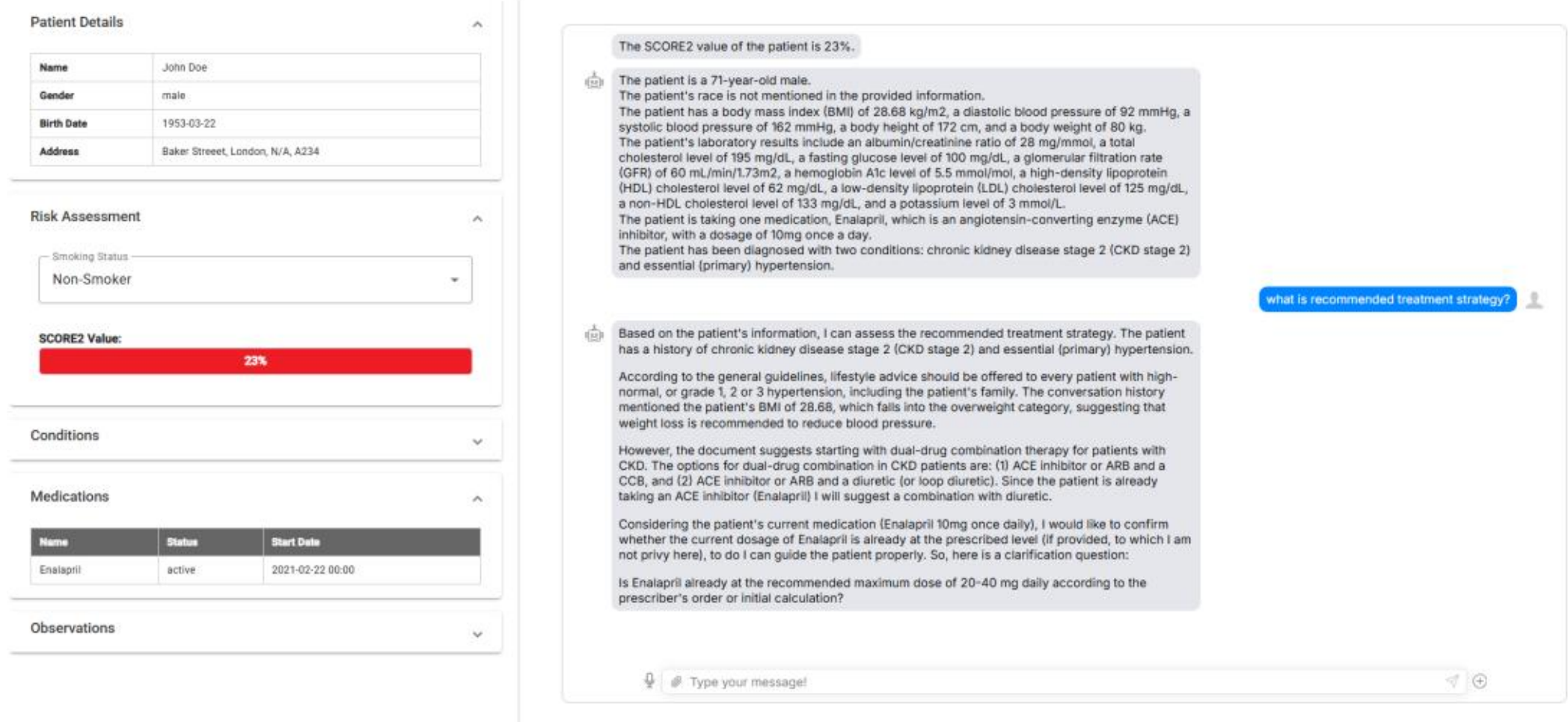

**Figure 2 –** User Interface implemented for RAG Execution

*2.3. Evaluation Framework*

Our evaluation starts with developing questions and their corresponding answers based on the guidelines provided. For this purpose, we tasked our evaluation panel of physicians to generate questions and answers related to their daily clinical practices. Questions ranged from simple ones to more complex scenario-based questions, e.g., "I have a patient with diabetes and hypertension. What should be the initial drug therapy?". In total, we compiled approximately 70 questions and answers, regarded as ground truth, which are made available as *Supplementary Material-5*.

After obtaining the answers to these questions from our RAG system, in the first step of the evaluation, we compared them with the ground truth answers using text similarity metrics Recall-Oriented Understudy for Gisting Evaluation (ROUGE) [30], Metric for Evaluation of Translation with Explicit Ordering (METEOR) [31] and BERTScore [32]. In text comparisons, BERTScore is used to evaluate semantic accuracy, while ROUGE/METEOR are used to assess word-level matching. Specifically:

- BERTScore can be the primary metric because, in healthcare, capturing the correct meaning of an answer is more important than exact word matches.
- ROUGE can be used to ensure that the model captures all necessary key terms from the guideline.
- METEOR can be a good middle-ground metric if flexibility is needed in language while still maintaining word overlap.

In this study, we opted not to use the BLEU (Bilingual Evaluation Understudy) [33]metric for evaluating the quality of model-generated medical summaries. Although BLEU is widely used for machine translation and other natural language generation tasks, it can be overly strict when applied to healthcare data. BLEU measures the degree of n-gram overlap between generated and reference text, which works well for tasks reliant on literal word matching. However, in healthcare, accurate interpretation and relevance of information are far more critical than exact phrasing. Medical summaries often vary in terminology, phrasing, or style while still conveying the same clinical meaning. The strictness of BLEU in penalizing even slight variations in wording can result in misleading evaluations of model performance in this context. Therefore, more appropriate metrics that focus on semantic accuracy and clinical relevance are preferred for evaluating health-related outputs.

These scores are developed to measure word overlap, sentence structure similarity, and semantic coherence but not factual correctness. For clinical questions, factual correctness is the most important feature. This is an important challenge that should be addressed, as current responses could appear lexically comparable to a reference answer but fail to capture the factual information necessary to guide clinical care. This can result in high scores for responses that are factually incorrect (false positives) or low scores for accurate responses that are phrased differently than the reference (false negatives). While useful for certain aspects of evaluation, these metrics fail to capture the nuances of medical relevance, completeness, and contextual correctness in the answers provided by the LLM. This limitation underscores the persistent need for expert physician oversight in the evaluation process, i.e., human-in-the-loop.

For automated grading of our RAG generated responses, we apply a recent approach called "LLM-as-a-Judge" [41], where LLM-evaluators are used to evaluate the quality of another LLM's response. For this approach, we employed two tools. The first one is Prometheus 2 [35], which is a finetuned evaluator (based on Llama 3.1 8B LLM) that performs fine-grained evaluation of text responses based on user-defined score rubrics. Prometheus 2 takes as input the instructions, score rubric, response to evaluate, and a gold reference answer, making it a referenced-based evaluator. Then, it scores the response to evaluate and returns text feedback. The prompt used by Prometheus 2 is as presented in Table 1.

**Table 1 –** The Prompt used by Prometheus 2

```
###Task Description:
```

```
An instruction (might include an Input inside it), a response to evaluate, a
reference answer that gets a score of 5, and a score rubric representing a
evaluation criteria are given.
1. Write detailed feedback that assesses the quality of the response strictly
based on the given score rubric, not evaluating in general.
2. After writing a feedback, write a score that is an integer between 1 and 5.
You should refer to the score rubric.
3. The output format should look as follows: "Feedback: {{write a feedback for
criteria}} [RESULT] {{an integer number between 1 and 5}}"
4. Please do not generate any other opening, closing, and explanations. Be sure
to include [RESULT] in your output.

###The instruction to evaluate:
{instruction}

###Response to evaluate:
{response}

###Reference Answer (Score 5):
{reference_answer}

###Score Rubrics:
[Is the response correct, accurate, and factual based on the reference answer?]
Score 1: The response is completely incorrect, inaccurate, and/or not factual.
Score 2: The response is mostly incorrect, inaccurate, and/or not factual.
Score 3: The response is somewhat correct, accurate, and/or factual.
Score 4: The response is mostly correct, accurate, and factual.
Score 5: The response is completely correct, accurate, and factual.
###Feedback:
```

The second LLM-evaluator tool is RAGAS [36] (Retrieval Augmented Generation Assessment), which is a framework evaluation of RAG pipelines by considering both the retriever and response generation phases. RAGAS offers metrics tailored for evaluating each component of a RAG pipeline in isolation. In our study, we used the metrics listed in Table 2.

**Table 2 – RAGAS metrics**

| Type | Metric | Description |
| --- | --- | --- |
| Retriever Metrics | Context Precision | In simple terms how relevant is context retrieved to the question asked. |
| | Context Recall | Is the retriever able to retrieve all the relevant context pertaining to ground truth? |
| Response | Answer Relevancy | How relevant is the generated answer to the question. |

| Generation Metrics | Faithfulness | Factual consistency of generated answers with the given context. |
|---|---|---|
| Comprehensive Metrics | Answer Correctness | Answer correctness encompasses two critical aspects: semantic similarity between the generated answer and the ground truth, as well as factual similarity. |
| | Answer Similarity | The semantic resemblance between the generated answer and the ground truth. |

Finally, we evaluated the performance of our RAG system against well-known medical LLMs (Meditron [9], BioMistral [10] and OpenBioLLM [11] and the original Llama 3.1 8B) by applying the above evaluation framework (except RAGAS which is only applicable to RAG systems) to these LLMs. All of these evaluation results including the questions asked to each system, answers received, the corresponding ground truth, the Prometheus evaluation results and evaluation scores from METEOR, BERTScore and ROUGE are made available as *Supplementary Material-4*. This folder also includes the results of RAGAS evaluation.

Importantly, we complemented this evaluation with a rigorous human assessment, where three independent physicians, leveraging their medical expertise, carefully evaluated the system's responses for accuracy, relevance, and clinical soundness. This human evaluation served as a critical validation step, ensuring that the system's outputs meet real-world medical expectations. The full list of evaluation questions and physicians' responses which are provided as *Supplementary Material-6*.

## 3. Results

Following the comprehensive evaluation framework outlined in the previous section, we assessed the performance of FHIR-RAG-MEDS against various large medical language models across four sections of the CAREPATH consensus clinical guidelines: dementia, COPD, hypertension, and sarcopenia. The evaluation aimed to determine the effectiveness of each model in generating accurate, relevant, and contextually appropriate responses to clinical queries. The results were measured using multiple metrics, including BERTScore, ROUGE-L, and METEOR, alongside a tailored set of RAGAS and Prometheus 2 metrics designed to evaluate the models' performance from a clinical decision-support perspective. This section presents the evaluation results, highlighting the strengths and weaknesses of each LLM in relation to the medical guidelines.

### *3.1. Dementia Results Interpretation*

For the dementia guideline, the proposed FHIR-RAG-MEDS system consistently outperformed the other models in all the text-based evaluation metrics (BERTScore F1, ROUGE-L F1, and METEOR), as shown in Table 3. Specifically, FHIR-RAG-MEDS achieved the highest Prometheus 2 average score of 4.0, indicating that its responses were judged to be more accurate, relevant, and clinically appropriate compared to the other evaluated LLMs. Among the baseline models, BioMistral achieved the second-highest Prometheus score of 3.5, while Meditron obtained the lowest score of 2.6667.

**Table 3 –** Comparison of FHIR-RAG-MEDS with other LLMs for the dementia guideline.

| LLM NAME | PROMETHEUS 2 AVERAGE SCORE | BERTSCORE F1 | ROUGE-L F1 | METEOR |
|---|---|---|---|---|

| **BIOMISTRAL** | 3.5000 | 0.5365 | 0.1418 | 0.1847 |
|---|---|---|---|---|
| **LLAMA 3.1 8B** | 3.1667 | 0.5851 | 0.1526 | 0.2800 |
| **MEDITRON** | 2.6667 | 0.5513 | 0.1394 | 0.2566 |
| **OPENBIOLLM** | 3.0000 | 0.5751 | 0.1531 | 0.2667 |
| **FHIR-RAG-MEDS** | 4.0000 | 0.6372 | 0.2543 | 0.3781 |

The quantitative text-based metrics further confirm the superior performance of FHIR-RAG-MEDS.

- The BERTScore F1 value of 0.6372 for FHIR-RAG-MEDS demonstrates strong semantic alignment between the generated responses and the reference answers, indicating high semantic accuracy. This score clearly exceeds those of the other models, with Llama 3.1 8B achieving 0.5851, OpenBioLLM 0.5751, Meditron 0.5513, and BioMistral 0.5365. These results confirm that FHIR-RAG-MEDS produced responses that were semantically closer to the guideline-based reference answers.
- The ROUGE-L F1 score of 0.2543 achieved by FHIR-RAG-MEDS highlights its superior ability to preserve key clinical concepts and important guideline terminology. This score is substantially higher than those of the other models, which ranged from 0.1394 (Meditron) to 0.1531 (OpenBioLLM). This indicates that the proposed system more effectively incorporated relevant guideline content into its responses.
- The METEOR score of 0.3781 further demonstrates the strong performance of FHIR-RAG-MEDS in generating linguistically flexible yet accurate responses. This score significantly surpasses those of Llama 3.1 8B (0.2800), OpenBioLLM (0.2667), Meditron (0.2566), and BioMistral (0.1847). This suggests that FHIR-RAG-MEDS was more effective in maintaining the intended meaning while allowing natural variations in wording.

In the RAGAS metrics for dementia, FHIR-RAG-MEDS demonstrates strong overall performance, as shown in Table 4. The answer similarity score of 0.942 is particularly high, indicating that the generated responses closely align semantically with the reference answers. Similarly, the context precision score of 0.903 confirms that the retrieved context is highly relevant and contains minimal irrelevant information. The answer correctness (0.780) and answer relevancy (0.806) scores further demonstrate that the system generates generally accurate and relevant responses. The context recall score of 0.776 indicates that most, though not all, relevant information was successfully retrieved. Finally, the faithfulness score of 0.859 shows a strong level of factual grounding in the retrieved context, confirming that the generated responses are largely supported by the source guideline content.

**Table 4 –** The RAGAS metrics of FHIR-RAG-MEDS for dementia, COPD, hypertension, and sarcopenia.

| **METRIC** | **DEMENTIA** | **COPD** | **Hypertension** | **Sarcopenia** |
|---|---|---|---|---|
| **ANSWER CORRECTNESS** | 0.780 | 0.882 | 0.952 | 0.878 |
| **ANSWER RELEVANCY** | 0.806 | 0.821 | 0.872 | 0.846 |
| **ANSWER SIMILARITY** | 0.942 | 0.929 | 0.908 | 0.939 |
| **CONTEXT PRECISION** | 0.903 | 0.962 | 0.958 | 1.000 |
| **CONTEXT RECALL** | 0.776 | 0.897 | 0.905 | 0.935 |
| **FAITHFULNESS** | 0.859 | 0.767 | 0.789 | 0.785 |

*3.2. COPD Results Interpretation*

For the COPD guideline, FHIR-RAG-MEDS continues to demonstrate superior performance, although the gap between models is narrower in some metrics compared to the dementia guideline. As shown in Table 5, the FHIR-RAG-MEDS system achieved the highest Prometheus 2 average score of 4.3846, confirming its strong ability to generate clinically relevant, accurate, and high-quality responses. Among the baseline models, OpenBioLLM (4.1538) and Llama 3.1 8B (4.0769) also performed well, while Meditron obtained the lowest score of 3.2308.

**Table 5 -** Comparison of FHIR-RAG-MEDS with other LLMs for the COPD guideline.

| LLM NAME | AVERAGE SCORE | BERTSCORE F1 | ROUGE-L F1 | METEOR |
|---|---|---|---|---|
| **BioMistral** | 3.8462 | 0.5624 | 0.1587 | 0.2140 |
| **Llama 3.1 8B** | 4.0769 | 0.5780 | 0.1557 | 0.3075 |
| **Meditron** | 3.2308 | 0.5623 | 0.1661 | 0.2727 |
| **OpenBioLLM** | 4.1538 | 0.5809 | 0.1800 | 0.2719 |
| **FHIR-RAG-MEDS** | 4.3846 | 0.6311 | 0.2585 | 0.3642 |

Table 5 presents the detailed evaluation results for the COPD guideline, which can be interpreted as follows:

- BERTScore F1: FHIR-RAG-MEDS achieved the highest score of 0.6311, indicating the strongest semantic alignment with the reference answers. The competing models showed slightly lower performance, with OpenBioLLM scoring 0.5809, Llama 3.1 8B 0.5780, Meditron 0.5623, and BioMistral 0.5624. These results confirm that FHIR-RAG-MEDS generated responses that were semantically closer to the guideline-based ground truth.
- ROUGE-L F1: FHIR-RAG-MEDS again achieved the best result with a score of 0.2585, demonstrating its superior ability to preserve important clinical terminology and key concepts from the COPD guideline. OpenBioLLM followed with 0.1800, while Meditron (0.1661), BioMistral (0.1587), and Llama 3.1 8B (0.1557) achieved lower scores. Although the gap is smaller than in the dementia guideline, FHIR-RAG-MEDS still clearly outperforms the other models.
- METEOR: FHIR-RAG-MEDS maintained its leading position with a score of 0.3642, reflecting its strong capability to generate responses that preserve meaning while allowing natural linguistic variation. Llama 3.1 8B achieved the second-highest score of 0.3075, followed by Meditron (0.2727), OpenBioLLM (0.2719), and BioMistral (0.2140). These results further confirm the advantage of the proposed system in producing flexible yet accurate responses.

For the RAGAS metrics in COPD, FHIR-RAG-MEDS demonstrates consistently strong performance across all evaluation dimensions, as shown in Table 4. The answer correctness score of 0.882 indicates a high level of factual accuracy in the generated responses. The answer similarity score of 0.929 further confirms strong semantic alignment with the reference answers. The context precision score of 0.962 is particularly high, indicating that the retrieved context is highly relevant and well-focused. Additionally, the context recall score of 0.897 demonstrates that most of the relevant information was successfully retrieved and utilized. The faithfulness score of 0.767, while slightly lower than some other metrics, still indicates good factual consistency between the generated responses and the supporting context. Overall, these results confirm that FHIR-RAG-MEDS produces reliable and clinically meaningful responses for the COPD guideline.

### *3.3. Hypertension Results Interpretation*

For the hypertension guideline, FHIR-RAG-MEDS further strengthens its leading position, particularly in terms of semantic accuracy and overall response quality. As shown in Table 6, the FHIR-RAG-MEDS system achieved the highest Prometheus 2 average score of 4.4474, demonstrating its superior effectiveness in generating accurate and clinically relevant responses. In comparison, the baseline models achieved significantly lower scores, with BioMistral obtaining 3.5135, Llama 3.1 8B 3.3514, OpenBioLLM 3.0811, and Meditron 2.9189.

**Table 6 -** Comparison of FHIR-RAG-MEDS with other LLMs for the hypertension guideline.

| LLM NAME | AVERAGE SCORE | BERTSCORE F1 | ROUGE-L F1 | METEOR |
|---|---|---|---|---|
| **BioMistral** | 3.5135 | 0.5258 | 0.1367 | 0.2121 |
| **Llama 3.1 8B** | 3.3514 | 0.5398 | 0.1140 | 0.2576 |
| **Meditron** | 2.9189 | 0.5145 | 0.1444 | 0.1857 |
| **OpenBioLLM** | 3.0811 | 0.5450 | 0.1432 | 0.2546 |
| **FHIR-RAG-MEDS** | 4.4474 | 0.6493 | 0.2986 | 0.4634 |

The numbers shown in Table 6 can be interpreted as follows.

- BERTScore F1: FHIR-RAG-MEDS achieved the highest score of 0.6493, indicating excellent semantic alignment between the generated responses and the reference answers. This score substantially exceeds those of OpenBioLLM (0.5450), Llama 3.1 8B (0.5398), BioMistral (0.5258), and Meditron (0.5145). These results confirm that FHIR-RAG-MEDS produced responses that most closely preserved the intended clinical meaning.
- ROUGE-L F1: With a score of 0.2986, FHIR-RAG-MEDS clearly outperformed all competing models, demonstrating its superior ability to capture and retain key clinical terminology and important guideline content. The other models achieved considerably lower scores, including Meditron (0.1444), OpenBioLLM (0.1432), BioMistral (0.1367), and Llama 3.1 8B (0.1140). This large margin highlights the effectiveness of the proposed system in incorporating relevant guideline information into its responses.
- METEOR: FHIR-RAG-MEDS again achieved the best performance with a score of 0.4634, reflecting its strong capability to generate accurate responses while maintaining flexibility in language expression. The competing models achieved notably lower scores, including Llama 3.1 8B (0.2576), OpenBioLLM (0.2546), BioMistral (0.2121), and Meditron (0.1857). This further confirms the robustness of FHIR-RAG-MEDS in preserving meaning while allowing natural linguistic variation.

In the RAGAS metrics for hypertension, FHIR-RAG-MEDS achieves its strongest overall performance among all evaluated guidelines, as shown in Table 4. The answer correctness score of 0.952 demonstrates excellent factual accuracy, indicating that the generated responses closely match the ground truth. The answer relevancy score of 0.872 and answer similarity score of 0.908 further confirm that the responses are both relevant and semantically aligned with the reference answers. The context precision (0.958) and context recall (0.905) scores indicate that the retrieval component is highly effective in identifying and utilizing the most relevant guideline content. The faithfulness score of 0.789 also demonstrates strong factual consistency between the generated responses and the retrieved context. These results confirm the robustness and reliability of FHIR-RAG-MEDS in the hypertension domain.

### *3.4. Sarcopenia Results Interpretation*

For the sarcopenia guideline, FHIR-RAG-MEDS is once again the top-performing model across all evaluation metrics. The system achieved the highest Prometheus 2 average score of 4.3636, confirming its superior ability to generate clinically accurate and relevant responses. The baseline models achieved lower scores, including Llama 3.1 8B (3.9091), Meditron (3.6364), BioMistral (3.5455), and OpenBioLLM (3.0909), highlighting the clear advantage of the proposed system.

**Table 7 -** Comparison of FHIR-RAG-MEDS with other LLMs for the sarcopenia guideline.

| LLM NAME | AVERAGE SCORE | BERTSCORE F1 | ROUGE-L F1 | METEOR |
|---|---|---|---|---|
| **BioMistral** | 3.5455 | 0.5901 | 0.1767 | 0.2460 |
| **Llama 3.1 8B** | 3.9091 | 0.5538 | 0.1230 | 0.2758 |
| **Meditron** | 3.6364 | 0.6132 | 0.1835 | 0.3305 |
| **OpenBioLLM** | 3.0909 | 0.6142 | 0.1968 | 0.3468 |
| **FHIR-RAG-MEDS** | 4.3636 | 0.7367 | 0.4648 | 0.6401 |

Table 7 lists the results of the evaluation for the sarcopenia guideline, which can be interpreted as follows.

- BERTScore F1: FHIR-RAG-MEDS achieved the highest score of 0.7367, indicating excellent semantic alignment between the generated responses and the reference answers. The closest competing models were OpenBioLLM (0.6142) and Meditron (0.6132), followed by BioMistral (0.5901) and Llama 3.1 8B (0.5538). These results demonstrate the superior capability of FHIR-RAG-MEDS in preserving the intended clinical meaning of the guideline content.
- ROUGE-L F1: With a score of 0.4648, FHIR-RAG-MEDS significantly outperformed all other models, more than doubling the performance of the closest competitor, OpenBioLLM (0.1968). The remaining models achieved lower scores, including Meditron (0.1835), BioMistral (0.1767), and Llama 3.1 8B (0.1230). This large performance gap highlights the effectiveness of FHIR-RAG-MEDS in accurately capturing and reproducing key clinical concepts and terminology from the sarcopenia guideline.

- METEOR: FHIR-RAG-MEDS achieved an outstanding score of 0.6401, demonstrating its strong ability to generate responses that maintain meaning while allowing natural variations in wording. The baseline models performed substantially worse, with OpenBioLLM scoring 0.3468, Meditron 0.3305, Llama 3.1 8B 0.2758, and BioMistral 0.2460. These results further confirm the robustness and linguistic flexibility of the proposed system.

The RAGAS metrics for sarcopenia further demonstrate the strong performance of FHIR-RAG-MEDS across all evaluation dimensions, as shown in Table 4. The answer similarity score of 0.939 indicates excellent semantic alignment with the reference answers. The answer correctness score of 0.878 and answer relevancy score of 0.846 confirm that the generated responses are both accurate and relevant. Notably, the context precision score reached 1.000, indicating that all retrieved context was fully relevant with no irrelevant information included. The context recall score of 0.935 further demonstrates that most of the relevant guideline information was successfully retrieved. The faithfulness score of 0.785 confirms good factual grounding in the retrieved context. Overall, these results demonstrate that FHIR-RAG-MEDS provides highly reliable and contextually accurate responses for the sarcopenia guideline.

Consequently, this breakdown provides a guideline-specific analysis, demonstrating FHIR-RAG-MEDS' overall superiority in generating factually accurate and semantically aligned responses across various medical guidelines. Each guideline presented unique challenges, but the FHIR-RAG-MEDS model proved to be the most robust, particularly in the sarcopenia and dementia domains.

High scores in BERTScore, ROUGE-L, and METEOR indicate strong semantic understanding and flexibility in language use, both of which are crucial for clinical decision-making. The RAGAS metrics further reinforce the reliability of the FHIR-RAG-MEDS system, showing that it effectively utilizes context and delivers factually correct responses. Although there is room for improvement in certain aspects of faithfulness and context recall, this comprehensive evaluation underscores FHIR-RAG-MEDS' capabilities, particularly in the medical domain, and its potential to support healthcare professionals in clinical decision-making.

### *3.5. Human Evaluation of FHIR-RAG-MEDS System*

In the context of clinical decision support systems, human evaluation remains a critical component of performance validation. While automated metrics such as BERTScore, ROUGE, and METEOR provide valuable insights into semantic and lexical alignment of generated responses, they often fall short in assessing factual correctness, clinical relevance, and contextual appropriateness, key factors in medical decision-making. Incorporating human evaluators, particularly domain experts like physicians, ensures a more comprehensive assessment of system outputs. These evaluations capture nuanced clinical considerations, identify potential gaps in reasoning, and validate the alignment of responses with evidence-based guidelines.

To evaluate the accuracy and clinical relevance of the FHIR-RAG-MEDS system, we conducted a human evaluation study involving three independent physicians, who are geriatricians and experts in multimorbidity in older adults with dementia. These physicians were tasked with reviewing and scoring the system's responses to questions derived from four clinical guidelines: dementia, COPD, hypertension, and sarcopenia. Responses were rated on a 5-point Likert scale from 1 (The response is completely incorrect, inaccurate, and/or not factual) to 5 (The response is completely correct, accurate, and factual). A standardized Excel template was developed to facilitate this process. Each physician independently assessed the system's responses, assigning numerical scores and providing qualitative comments where applicable. These evaluations aimed to measure the factual correctness, clarity, clinical relevance, and adherence to the respective clinical guidelines.

Across the four guidelines, the average physician scores ranged from 3.67 to 4.45, with the highest scores observed for hypertension-related queries and the lowest for dementia-related queries. Specifically:

- COPD: The average physician score was 4.38 (±0.12), indicating strong agreement regarding the clinical relevance and accuracy of the responses.
- Dementia: The average score was 3.67 (±0.15), with variability reflecting the complexity of the scenarios and occasional gaps in contextual understanding.
- Hypertension: The system achieved the highest average score of 4.45 (±0.10), demonstrating both high consistency and reliability in this domain.
- Sarcopenia: The average score was 4.36 (±0.14), with qualitative feedback highlighting areas for improvement, particularly in providing actionable recommendations.

Physicians' qualitative feedback identified key strengths, such as the system's ability to generate contextually relevant, evidence-based recommendations. However, feedback also revealed occasional shortcomings, notably in capturing nuanced clinical details. For instance, 12% of responses were marked as "lacking specific actionable insights," despite being semantically accurate. The full list of questions and physicians' responses are provided as *Supplementary Material-6*.

Statistical analysis of inter-rater reliability using Cohen's kappa demonstrated substantial agreement among the three physicians ($\kappa = 0.79$), indicating a high degree of consistency in their evaluations. Additionally, the correlation between automated scores and physician scores was strong (Pearson's $r = 0.85$, $p < 0.01$), reinforcing the validity of the automated evaluation framework as a reliable complementary tool.

By combining human evaluation with automated metrics, this study highlights the critical role of expert oversight in validating AI-driven clinical decision support systems. This integrated approach ensures that the recommendations generated by the FHIR-RAG-MEDS system are not only technically robust but also clinically meaningful and actionable.

## 5. Discussions

### *5.1. Principal Findings*

Integrating LLMs into clinical decision support systems (CDSS) represents a significant shift in healthcare delivery, particularly by leveraging natural language processing to interpret clinical documentation and provide recommendations aligned with current medical research and best practices [44,45]. Our approach demonstrates that a locally hosted LLM, securely deployed in a private network, can access patient-specific data and integrate this information into clinical decision support. This enables the generation of highly personalized treatment plans based on up-to-date clinical guidelines and tailored to the unique needs of individual patients.

The LLMs tested, including BioMistral, Llama 3.1 8B, Meditron 3, and OpenBioLLM, showed varying degrees of accuracy in aligning with clinical guidelines. Notably, our RAG system, FHIR-RAG-MEDS, outperformed traditional LLMs in many scenarios, emphasizing the value of combining retrieval and generation techniques for more reliable and contextually accurate medical advice. The system's high performance across automated and human evaluations validates its ability to generate accurate and clinically relevant recommendations.

The integration of HL7 FHIR with RAG systems holds transformative potential for clinical decision support. By leveraging real-time access to patient data and combining it with up-to-date medical knowledge, this approach provides healthcare professionals with powerful tools to deliver personalized, evidence-based care. Integrating SMART on FHIR into the FHIR-RAG-MEDS system significantly enhances its interoperability, making it a versatile and scalable solution for clinical decision support. This implementation not only ensures compliance with modern healthcare standards but also positions the system as a robust tool capable of addressing diverse clinical scenarios through seamless connectivity with FHIR servers.

### *5.2. Limitations and Future Work*

Despite its contributions, this work is subject to limitations from both clinical and technical perspectives. First, from a clinical perspective, this study used 70 case scenarios created as question/answer pairs by expert evaluation panel representing clinical cases, rather than real clinical cases. This strategy inevitably constrains generalizability and external validation; however, assessing LLM performance using a deliberately constructed set of heterogeneous clinical scenarios offers greater analytical value than relying on general population samples that rarely capture uncommon or challenging cases [26]. As a future step, we aim to validate the robustness of FHIR-RAG-MEDS system against retrospective and prospective real-world clinical cases.

From a technical perspective, to further optimize the performance of LLMs within CDSS, future work could focus on improving accuracy, efficiency, and user experience. Accuracy can be enhanced by refining the system's ability to interpret complex clinical cases, ensuring that model outputs are both factually correct and clinically relevant. A key future direction is continual learning, whereby the system updates itself with new medical research and guidelines in real time, keeping pace with the rapidly evolving healthcare landscape. By integrating more robust human-in-the-loop feedback mechanisms, the system can evolve alongside medical professionals, fostering a collaborative environment where LLM-driven recommendations can be continually refined through expert feedback.

Extending the system with reinforcement learning with human feedback (RLHF), a technique that improves AI models' performance by learning directly from human feedback during the fine-tuning phase, could enhance accuracy and reduce bias [46]. In this approach, feedback provided by physicians can be used for reward modelling, which is then employed to fine-tune the model. This process could lead to the system generating responses that better align with human values and preferences. With the involvement of more physicians, and hence more expert feedback, RLHF could provide significant improvement in the system's performance.

Finally, as part of our future work, we aim to enhance the explainability of the decision support system by providing clear and transparent links to relevant sections of the evidence-based guidelines used in generating the recommendations. This focus on explainability will empower healthcare professionals to better understand the rationale behind the system's suggestions, fostering trust and facilitating informed decision-making. By integrating explicit references to the specific guidelines or evidence sources that underpin the personalized advice, the system will not only support clinical accuracy but also ensure accountability and reliability in its outputs. This approach aligns with the principles of interpretable AI and is expected to improve user confidence and adoption.

## 6. Conclusions

This study demonstrates that integrating retrieval-augmented generation with HL7 FHIR enables trustworthy, patient-specific, and guideline-concordant clinical decision support. By combining real-time access to structured EHR data with evidence-based clinical guidelines, FHIR-RAG-MEDS overcomes key limitations of standalone medical LLMs related to personalization, transparency, and clinical reliability. The results show that seamless interoperability through standardized frameworks such as SMART on FHIR is essential for translating advances in generative AI into clinically meaningful decision support. FHIR-RAG-MEDS provides a scalable and interoperable foundation for next-generation clinical decision support systems that align with real-world clinical workflows and patient-centered care.

## Abbreviations

The following abbreviations are used in this manuscript:

| | |
|---|---|
| AI | Artificial Intelligence |
| BLEU | Bilingual Evaluation Understudy |
| CDSS | Clinical Decision Support Systems |
| COPD | Chronic Obstructive Pulmonary Disease |
| CRG | Clinical Reference Group |
| EC | European Commission |
| EHR | Electronic Health Records |
| FHIR | Fast Healthcare Interoperability Resources |
| HL7 | Health Level 7 |
| LLM | Large Language Model |
| MCI | Mild Cognitive Impairment |
| MD | Mild Dementia |
| METEOR | Metric for Evaluation of Translation with Explicit Ordering |
| RAG | Retrieval-Augmented Generation |
| RAGAS | Retrieval Augmented Generation Assessment |
| RLHF | Reinforcement Learning with Human Feedback |
| ROUGE | Recall-Oriented Understudy for Gisting Evaluation |

## References

1. Lugtenberg, M.; Burgers, J.S.; Westert, G.P. Effects of Evidence-Based Clinical Practice Guidelines on Quality of Care: A Systematic Review. *Qual. Saf. Health Care* **2009**, *18*, 385–392, doi:10.1136/qshc.2008.028043.
2. Lichtner, G.; Spies, C.; Jurth, C.; Bienert, T.; Mueller, A.; Kumpf, O.; Piechotta, V.; Skoetz, N.; Nothacker, M.; Boeker, M.; et al. Automated Monitoring of Adherence to Evidenced-Based Clinical Guideline Recommendations: Design and Implementation Study. *J. Med. Internet Res.* **2023**, *25*, e41177, doi:10.2196/41177.
3. Fischer, F.; Lange, K.; Klose, K.; Greiner, W.; Kraemer, A. Barriers and Strategies in Guideline Implementation—A Scoping Review. *Healthcare* **2016**, *4*, 36, doi:10.3390/healthcare4030036.
4. Riaño, D.; Peleg, M.; ten Teije, A. Ten Years of Knowledge Representation for Health Care (2009–2018): Topics, Trends, and Challenges. *Artif. Intell. Med.* **2019**, *100*, 101713, doi:10.1016/j.artmed.2019.101713.
5. Laleci Erturkmen, G.B.; Yuksel, M.; Sarigul, B.; Arvanitis, T.N.; Lindman, P.; Chen, R.; Zhao, L.; Sadou, E.; Bouaud, J.; Traore, L.; et al. A Collaborative Platform for Management of Chronic Diseases via Guideline-Driven Individualized Care Plans. *Comput. Struct. Biotechnol. J.* **2019**, *17*, 869–885, doi:10.1016/j.csbj.2019.06.003.
6. García-Lorenzo, B.; Gorostiza, A.; González, N.; Larrañaga, I.; Mateo-Abad, M.; Ortega-Gil, A.; Bloemeke, J.; Groene, O.; Vergara, I.; Mar, J.; et al. Assessment of the Effectiveness, Socio-Economic Impact and Implementation of a Digital Solution for Patients with Advanced Chronic Diseases: The ADLIFE Study Protocol. *Int. J. Environ. Res. Public Health* **2023**, *20*, 3152, doi:10.3390/ijerph20043152.

7. Ulgu, M.M.; Laleci Erturkmen, G.B.; Yuksel, M.; Namli, T.; Postacı, Ş.; Gencturk, M.; Kabak, Y.; Sinaci, A.A.; Gonul, S.; Dogac, A.; et al. A Nationwide Chronic Disease Management Solution via Clinical Decision Support Services: Software Development and Real-Life Implementation Report. *JMIR Med. Inform.* **2024**, *12*, e49986, doi:10.2196/49986.
8. Gencturk, M.; Laleci Erturkmen, G.B.; Akpinar, A.E.; Pournik, O.; Ahmad, B.; Arvanitis, T.N.; Schmidt-Barzynski, W.; Robbins, T.; Alcantud Corcoles, R.; Abizanda, P. Transforming Evidence-Based Clinical Guidelines into Implementable Clinical Decision Support Services: The CAREPATH Study for Multimorbidity Management. *Front. Med. (Lausanne).* **2024**, *11*, doi:10.3389/fmed.2024.1386689.
9. Zeming Chen, A.H.C. et. al. MEDITRON-70B: Scaling Medical Pretraining for Large Language Models. *ArXiv* **2023**, *abs/2311.16079*.
10. Yanis Labrak, A.B.E.M.P.-A.G.M.R.R.D. BioMistral: A Collection of Open-Source Pretrained Large Language Models for Medical Domains. In Proceedings of the ACL 2024 - Proceedings of the 62st Annual Meeting of the Association for Computational Linguistics; 2024.
11. Open Source Biomedical Large Language Model 2024.
12. Patrick Lewis, E.P.A.P.F.P.V.K.N.G.H.K.M.L.W.Y.T.R.S.R.D.K. Retrieval-Augmented Generation for Knowledge-Intensive NLP Tasks. *ArXiv* **2021**, doi:https://doi.org/10.48550/arXiv.2005.11401.
13. Miao, J.; Thongprayoon, C.; Suppadungsuk, S.; Garcia Valencia, O.A.; Cheungpasitporn, W. Integrating Retrieval-Augmented Generation with Large Language Models in Nephrology: Advancing Practical Applications. *Medicina (B Aires).* **2024**, *60*, 445, doi:10.3390/medicina60030445.
14. Tian, S.; Jin, Q.; Yeganova, L.; Lai, P.-T.; Zhu, Q.; Chen, X.; Yang, Y.; Chen, Q.; Kim, W.; Comeau, D.C.; et al. Opportunities and Challenges for ChatGPT and Large Language Models in Biomedicine and Health. *Brief. Bioinform.* **2023**, *25*, doi:10.1093/bib/bbad493.
15. Ji, Z.; Lee, N.; Frieske, R.; Yu, T.; Su, D.; Xu, Y.; Ishii, E.; Bang, Y.J.; Madotto, A.; Fung, P. Survey of Hallucination in Natural Language Generation. *ACM Comput. Surv.* **2023**, *55*, 1–38, doi:10.1145/3571730.
16. Yunfan Gao, Y.X.X.G.K.J.J.P.Y.B.Y.D.J.S.M.W.H.W. Retrieval-Augmented Generation for Large Language Models: A Survey. *ArXiv* **2024**, doi:https://doi.org/10.48550/arXiv.2312.10997.
17. Penghao Zhao, H.Z.Q.Y.Z.W.Y.G.F.F.L.Y.W.Z.J.J.B.C. Retrieval-Augmented Generation for AI-Generated Content: A Survey. *ArXiv* **2024**, doi:https://doi.org/10.48550/arXiv.2402.19473.
18. Junde Wu, J.Z.Y.Q. Medical Graph RAG: Towards Safe Medical Large Language Model via Graph Retrieval-Augmented Generation. *ArXiv* **2024**, doi:https://doi.org/10.48550/arXiv.2408.04187.
19. Zakka, C.; Shad, R.; Chaurasia, A.; Dalal, A.R.; Kim, J.L.; Moor, M.; Fong, R.; Phillips, C.; Alexander, K.; Ashley, E.; et al. Almanac — Retrieval-Augmented Language Models for Clinical Medicine. *NEJM AI* **2024**, *1*, doi:10.1056/AIoa2300068.
20. YuHe Ke, L.J.K.E.H.R.A.N.L.A.T.H.S.C.R.S.J.Y.M.T.J.C.L.O.D.S.W.T. Development and Testing of Retrieval Augmented Generation in Large Language Models -- A Case Study Report. *ArXiv* **2024**, doi:https://doi.org/10.48550/arXiv.2402.01733.
21. Nazi, Z. Al; Peng, W. Large Language Models in Healthcare and Medical Domain: A Review. *Informatics* **2024**, *11*, 57, doi:10.3390/informatics11030057.
22. Lingyao Li, J.Z.Z.G.W.H.L.F.H.Y.L.H.Y.Z.T.L.A.L.H.S.M. A Scoping Review of Using Large Language Models (LLMs) to Investigate Electronic Health Records (EHRs). *ArXiv* **2024**, doi:https://doi.org/10.48550/arXiv.2405.03066.
23. Paul Schmiedmayer, A.R.P.Z.V.R.A.Z.A.F.O.A. LLM on FHIR -- Demystifying Health Records. *ArXiv* **2024**, doi:https://doi.org/10.48550/arXiv.2402.01711.
24. Li, Y.; Wang, H.; Yerebakan, H.Z.; Shinagawa, Y.; Luo, Y. FHIR-GPT Enhances Health Interoperability with Large Language Models. *NEJM AI* **2024**, *1*, doi:10.1056/AIcs2300301.

25. Lewis, M.; Thio, S.; Roberts, A.; Siju, C.; Mukit, W.; Kuruvilla, R.; Jiang, Z.J.; Möller-Grell, N.; Borakati, A.; Dobson, R.J.; et al. *Grounding Large Language Models in Clinical Evidence: A Retrieval-Augmented Generation System for Querying UK NICE Clinical Guidelines*; 2025;
26. Tung, J.Y.M.; Le, Q.; Yao, J.; Huang, Y.; Lim, D.Y.Z.; Sng, G.G.R.; Lau, R.S.E.; Tan, Y.G.; Chen, K.; Tay, K.J.; et al. Performance of Retrieval-Augmented Generation Large Language Models in Guideline-Concordant Prostate-Specific Antigen Testing: Comparative Study With Junior Clinicians. *J. Med. Internet Res.* **2025**, *27*, e78393–e78393, doi:10.2196/78393.
27. Alkhalaf, M.; Yu, P.; Yin, M.; Deng, C. Applying Generative AI with Retrieval Augmented Generation to Summarize and Extract Key Clinical Information from Electronic Health Records. *J. Biomed. Inform.* **2024**, *156*, 104662, doi:10.1016/j.jbi.2024.104662.
28. Kresevic, S.; Giuffrè, M.; Ajcevic, M.; Accardo, A.; Crocè, L.S.; Shung, D.L. Optimization of Hepatological Clinical Guidelines Interpretation by Large Language Models: A Retrieval Augmented Generation-Based Framework. *NPJ Digit. Med.* **2024**, *7*, 102, doi:10.1038/s41746-024-01091-y.
29. Unlu, O.; Shin, J.; Mailly, C.J.; Oates, M.F.; Tucci, M.R.; Varugheese, M.; Wagholikar, K.; Wang, F.; Scirica, B.M.; Blood, A.J.; et al. Retrieval-Augmented Generation–Enabled GPT-4 for Clinical Trial Screening. *NEJM AI* **2024**, *1*, doi:10.1056/AIoa2400181.
30. Xiong, G.; Jin, Q.; Lu, Z.; Zhang, A. Benchmarking Retrieval-Augmented Generation for Medicine. **2024**.
31. FHIR-RAG-MED Interpretation System Available online: https://github.com/srdc/fhir-rag-med-interpret (accessed on 24 February 2026).
32. CAREPATH Project Website Available online: https://cordis.europa.eu/project/id/945169 (accessed on 24 February 2026).
33. Robbins, T.D.; Muthalagappan, D.; O'Connell, B.; Bhullar, J.; Hunt, L.-J.; Kyrou, I.; Arvanitis, T.N.; Keung, S.N.L.C.; Muir, H.; Pournik, O.; et al. Protocol for Creating a Single, Holistic and Digitally Implementable Consensus Clinical Guideline for Multiple Multi-Morbid Conditions. In Proceedings of the Proceedings of the 10th International Conference on Software Development and Technologies for Enhancing Accessibility and Fighting Info-exclusion; ACM: New York, NY, USA, August 31 2022; pp. 1–6.
34. LangChain Open Source Library 2026.
35. Chroma Open Source AI Application Database 2026.
36. Llama 3.1 8B LLM 2026.
37. Chin-Yew Lin ROUGE: A Package for Automatic Evaluation of Summaries. In Proceedings of the In Proceedings of the Workshop on Text Summarization Branches Out (WAS 2004); 2004.
38. Satanjeev Banerjee, A.L. METEOR: An Automatic Metric for MT Evaluation with Improved Correlation with Human Judgments. In Proceedings of the Proceedings of the ACL Workshop on Intrinsic and Extrinsic Evaluation Measures for Machine Translation and/or Summarization; 2005.
39. Tianyi Zhang, V.K.F.W.K.Q.W.Y.A. BERTScore: Evaluating Text Generation with BERT. In Proceedings of the International Conference on Learning Representations; 2020.
40. Papineni, K.; Roukos, S.; Ward, T.; Zhu, W.-J. BLEU. In Proceedings of the Proceedings of the 40th Annual Meeting on Association for Computational Linguistics - ACL '02; Association for Computational Linguistics: Morristown, NJ, USA, 2001; p. 311.
41. Yan, Z. Evaluating the Effectiveness of LLM-Evaluators (Aka LLM-as-Judge) Available online: https://eugeneyan.com/writing/llm-evaluators/ (accessed on 24 February 2026).
42. Seungone Kim, J.S.S.L.B.Y.L.J.S.S.W.G.N.M.L.K.L.M.S. Prometheus 2: An Open Source Language Model Specialized in Evaluating Other Language Models. *ArXiv* **2024**, doi:https://doi.org/10.48550/arXiv.2405.01535.
43. RAGAS Evaluation Framework 2026.

44. Mahadevaiah, G.; RV, P.; Bermejo, I.; Jaffray, D.; Dekker, A.; Wee, L. Artificial Intelligence-based Clinical Decision Support in Modern Medical Physics: Selection, Acceptance, Commissioning, and Quality Assurance. *Med. Phys.* **2020**, *47*, doi:10.1002/mp.13562.
45. Golden, G.; Popescu, C.; Israel, S.; Perlman, K.; Armstrong, C.; Fratila, R.; Tanguay-Sela, M.; Benrimoh, D. Applying Artificial Intelligence to Clinical Decision Support in Mental Health: What Have We Learned? *Health Policy Technol.* **2024**, *13*, 100844, doi:10.1016/j.hlpt.2024.100844.
46. Chaudhari, S.; Aggarwal, P.; Murahari, V.; Rajpurohit, T.; Kalyan, A.; Narasimhan, K.; Deshpande, A.; Castro da Silva, B. RLHF Deciphered: A Critical Analysis of Reinforcement Learning from Human Feedback for LLMs. *ACM Comput. Surv.* **2026**, *58*, 1–37, doi:10.1145/3743127.